\documentclass[conference]{IEEEtran}
\IEEEoverridecommandlockouts

\usepackage{cite}
\usepackage{amsmath,amssymb,amsfonts}
\usepackage{algorithmic}
\usepackage{graphicx}
\usepackage{textcomp}
\usepackage{xcolor}
\usepackage{pifont}
\def\BibTeX{{\rm B\kern-.05em{\sc i\kern-.025em b}\kern-.08em
    T\kern-.1667em\lower.7ex\hbox{E}\kern-.125emX}}

\makeatletter
\newcommand{\linebreakand}{%
  \end{@IEEEauthorhalign}
  \hfill\mbox{}\par
  \mbox{}\hfill\begin{@IEEEauthorhalign}
}
\makeatother

\newcommand{\ourmethod}{Traffic-Scene-Graph Inference}
\newcommand{\abbv}{TSGi}

\usepackage{booktabs}
\usepackage{float}
    
\begin{document}

\title{Enhancing Vision-Language Models with Scene Graphs for Traffic Accident Understanding}


\author{\IEEEauthorblockN{Aaron Lohner\textsuperscript{\ding{41}}$^{*}$\thanks{\textsuperscript{\ding{41}}Correspondence. *Work done while at CMU. **Equal advising.}}
\IEEEauthorblockA{\textit{Carnegie Mellon University}\\
Pittsburgh, USA \\
{\tt\small alohner@alumni.cmu.edu}}
\and
\IEEEauthorblockN{Francesco Compagno}
\IEEEauthorblockA{\textit{University of Trento}\\
Trento, Italy \\
{\tt\small francesco.compagno@unitn.it}}
\linebreakand
\IEEEauthorblockN{Jonathan Francis\textsuperscript{\ding{41}}$^{**}$}
\IEEEauthorblockA{\textit{Bosch Center for Artificial Intelligence;} \\
\textit{Carnegie Mellon University}\\
Pittsburgh, USA\\
{\tt\small jon.francis@us.bosch.com}}
\and
\IEEEauthorblockN{Alessandro Oltramari$^{**}$}
\IEEEauthorblockA{\textit{Bosch Center for Artificial Intelligence;} \\
\textit{Carnegie Mellon University}\\
Pittsburgh, USA \\
{\tt\small alessandro.oltramari@us.bosch.com}}
}

\maketitle

\begin{abstract}
Recognizing a traffic accident is an essential part of any autonomous driving or road monitoring system. An accident can appear in a wide variety of forms, and understanding what type of accident is taking place may be useful to prevent it from recurring. This work focuses on classifying traffic scenes into specific accident types. We approach the problem by representing a traffic scene as a graph, where objects such as cars can be represented as nodes, and relative distances and directions between them as edges. This representation of a traffic scene is referred to as a \textit{scene graph}, and can be used as input for an accident classifier. Better results are obtained with a classifier that fuses the scene graph input with visual and textual representations. This work introduces a multi-stage, multimodal pipeline that pre-processes videos of traffic accidents, encodes them as scene graphs, and aligns this representation with vision and language modalities before executing the classification task. When trained on 4 classes, our method achieves a balanced accuracy score of 57.77\% on an (unbalanced) subset of the popular Detection of Traffic Anomaly (DoTA) benchmark, representing an increase of close to 5 percentage points from the case where scene graph information is not taken into account.
\end{abstract}


\section{Introduction}
The task of understanding traffic scenarios is one with ever-increasing importance, particularly for the advancement of Autonomous Vehicles (AV) and road infrastructure systems \cite{xu2021sutdtrafficqa, dota, qasemi2023traffic,francis2022distribution, huang2024cadre, stoler2024seal, stoler2024safeshift}. A major aspect of this task is to efficiently and accurately recognize different types of traffic accidents, with the ultimate goal of preventing them. We approach this challenge by modeling traffic scenes using scene graphs. To improve classification performance, the graph representation is further aligned with representations from the vision and language modalities within contrastively-trained foundation models. Starting with video clips of traffic incidents, procured from the \textit{Detection of Traffic Anomaly} (DoTA) dataset \cite{dota}, we build \ourmethod~(\abbv), a multi-stage, multimodal pipeline that pre-processes videos, encodes them as scene graphs, and aligns these representations with vision and language modalities before classifying traffic accidents.

We propose \abbv{} as a unified system for accident classification.
To start, in \abbv{} we generate scene graphs using the \texttt{roadscene2vec} (rs2v) tool \cite{rs2v}. Then, we use a scene graph encoder (SGE) from the same authors \cite{spatio_temporal_sge} of rs2v to obtain encodings that are aligned to encodings of textual and visual inputs, obtained from CLIP \cite{clip} and X-CLIP \cite{xclip}, respectively\footnote{CLIP is a multi-modal model that is capable of zero-shot image classification by measuring text-image similarity, and its main components are a text encoder and an image encoder. X-CLIP expands upon CLIP's image encoder to include an attention mechanism to model inter-frame communication and to generate a new embedding representation from video frames.}. 
In essence, scene graphs are treated as a new modality (or `view') that is aligned with text and video signals, and are all used together to classify traffic accident scenes. 

A summary of contributions is as follows:


\begin{itemize}
    \item We introduce \ourmethod~(\abbv), a novel method for traffic accident classification that leverages scene graphs to capture essential features from traffic accidents.
    \item We show that the added signal from a scene graph modality can enhance the performance of a video-language traffic accident classifier by nearly 5 percentage points.
    \item Experiments in this work demonstrate that aligning the scene graph modality with vision and language gives similar results to omitting alignment training, although increasing the batch size and training time during alignment shows a trend of increasing scores and the potential in further improving classification results.
\end{itemize}

\begin{figure*}[htbp]
    \centering    \includegraphics[width=0.95\linewidth]{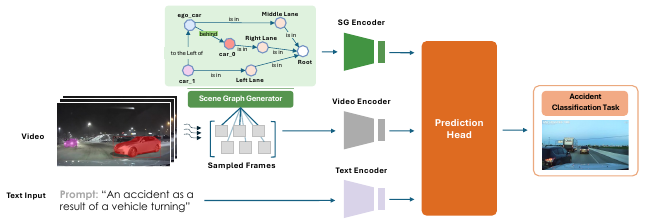}
    \caption{An overview of the \abbv~Architecture. Beginning with video and text inputs, video frames are sampled and used to generate scene graphs. Then, alignment training is performed on the three encoders before a prediction head is used to classify the accident.}
    \label{fig:arch}
\end{figure*}

\section{Related Work}
\subsection{Scene Graphs for Representing Traffic Scenes}
Several works have taken the approach of modeling traffic scenarios using scene graphs (SGs). One such representation proposed in \cite{yu2020scenegraph} forms the basis of the SGs produced in our work. The authors define a graph structure with nodes representing entities such as vehicles and pedestrians while edges represent relations between the objects and are labeled by distance category (e.g., \textit{visible, very near}) and orientation (e.g., \textit{in front of, beside}). Objects in a traffic scene are also modeled in \cite{semsgg}, further accounting for relative speeds of objects in the graph edges. In \cite{kgg}, semantic information is extracted from a traffic scene as well, but the focus is largely on generating graphs from road signs and lanes to understand how a traffic scene may be interpreted from the perspective of a driver.
For the related task of Question Answering in the domain of traffic accidents, \cite{qasemi2023traffic} uses a knowledge graph to enhance a video-language model. The method is to inject domain-specific knowledge by first auto-generating video annotations and storing semantic information about a traffic scene in a knowledge graph. It is interesting to contrast such an approach, which uses a knowledge graph to help contextualize a traffic scene in a broader traffic monitoring ontology, with our scene graph method, which builds a unique graphical structure from each scene. In the former case, the knowledge graph can answer general traffic accident-related questions, such as the impact of having more vehicles in a traffic scene, but is less effective in identifying accident types based on the interactions among vehicles in a specific traffic scene. Instead, our method is better suited for the task of accident classification by leveraging scene graphs to capture relationships among objects that may be involved in a traffic accident in a given video example.
Combining the use of scene graphs with knowledge graphs to model traffic settings, \cite{sgg_in_ad} makes use of a knowledge graph to boost the performance of the RelTR (Relation Transformer) model \cite{reltr}, an encoder-decoder model used for scene graph generation by framing it as a set prediction problem.


\subsection{Leveraging Scene Graphs with Neural Networks}
There are many possible designs and applications for leveraging the versatile structure of scene graphs to enhance the performance of neural networks \cite{regat, DG-PGNN, sayplan, saxena2024grapheqa}. As discussed in \cite{leveragingsgs}, Visual Question Answering (VQA) systems may be enhanced using scene graphs as an alternative to solely focusing on vision and language features. Although their study is not specific to our traffic domain, \cite{leveragingsgs} claims that scene graphs derived from images can capture the essential visual features and may outperform images for VQA tasks. Many works have focused on automatically generating scene graphs that can easily be encoded into neural networks for various downstream tasks. For example, the rs2v tool \cite{rs2v} generates scene graphs for individual frames of traffic videos and is applied to our work. Alternatively, \cite{nag2023unbiased} designs a tool specifically for modeling videos with scene graphs which accounts for spatial and temporal relationships between frames based on sequence modeling with transformers. However, they train on the Action Genome dataset \cite{ag} and produce graphs that are unrelated to traffic scenarios, making this tool less applicable to our work.

\subsection{Multimodal Alignment with Contrastive Representations}
In \cite{clip}, a model is proposed that uses a contrastive training strategy to align text captions with images for the Contrastive Language–Image Pre-training (CLIP) model. From this, several works have focused on aligning encoders of additional modalities to those trained for CLIP, such as audio \cite{wu2022wav2clip} and haptic modalities \cite{tatiya2024mosaic}. Similar to our implementation, \cite{koch2023lang3dsg} uses CLIP's text encoder to contrastively train 3D scene graphs, which are graphs that implicitly contain spatial information of objects in a scene, and can also model other relationships between these objects \cite{koch2023lang3dsg}. The work in \cite{huang2023structureclip} aims to enhance CLIP's performance by injecting scene graph information into the embedding generated by CLIP's text encoder before contrastive learning is performed between the image and text embedding spaces. To the best of our knowledge, however, our work is the first to treat scene graphs as an additional modality that is grounded to a text and video encoder. There are also many possible strategies for fusing the outputs of a multimodal model before the downstream task at hand, such as early and late stage fusion involving concatenation \cite{mmclftn, mmanal}, merging \cite{mmclftn}, or sampling from a shared embedding space \cite{mmsurvey}. We experiment with several of these late fusion techniques on the three modality embeddings after training for alignment.

\section{Methodology}

To approach classification of traffic accidents, a multi-stage pipeline was developed. The pipeline, which is represented in Figure \ref{fig:arch}, can be summarized in four stages that are described in the following sections:
\begin{enumerate}
    \item \textbf{Data pre-processing}: Sample video frames, generate captions, and use a scene graph generator to produce a set of scene graphs for each traffic accident example.
    \item \textbf{Scene graph encoder pre-training:} Pre-train the scene graph encoder on the classification task.
    \item \textbf{Multimodal alignment:} Align the scene graph encoder with frozen video and text encoders.
    \item \textbf{Fine tuning for downstream task:} Train a classification head on top of the aligned multimodal model.
\end{enumerate}



\subsection{Data Pre-processing}
In the first stage, for extensibility and ease, we leverage and tune existing tools for generating traffic scene graphs that will later be fed into our modeling approach. Various scene graph generators (SGGs) are available for this, although not many fit our specific requirements for a generator that is capable of identifying and encoding features unique to traffic accidents, such as different types of vehicle collisions. Some do work for our needs — in particular, the \texttt{roadscene2vec} (rs2v) tool \cite{rs2v} that we leverage for generating scene graphs from traffic video frames. To use rs2v, we first sample a fixed amount of frames from each video, and for each class we manually generate text captions.
Then, feeding in a series of frames from a traffic scene, the rs2v generator uses an object detector and keeps only the relevant entities relating to traffic (such as road, cars, pedestrians, etc.), filtering out other detected objects. Next, after generating a ``bird's eye view'' (BEV) projection of the image and approximating the relative location of each object in this projection, edges are added to connect nearby entities in the scene and to map vehicles to lanes. This forms the scene graph for a specific frame from a video sample. The SGG is a crucial component of this pipeline as it defines the elements in the scene graph modality.

An important step before employing this tool is to calibrate the BEV and adjust the proximity thresholds (which are used when creating edges labeled with varying attributes, such as \textit{very near} or \textit{visible}, relating a pair of objects). The authors specifically note the challenges they faced when adjusting these settings with regard to the DoTA dataset, and mention that for best results, they needed to calibrate their model for each traffic scene \cite{spatio_temporal_sge}. In our case, we generate SGs for a random subset of DoTA traffic scenes, visually inspect them for errors (e.g., the SGG labels the edge relating the ego vehicle to a vehicle directly in front of it as \textit{visible} rather than \textit{very near}), and then manually adjust the BEV parameters and proximity thresholds in an iterative manner. We do this to select one configuration to generate the SGs for our task, although it remains an inherent challenge to generalize the parameter settings for the entire dataset. Some examples of the generated SGs can be seen in Appendix \ref{appendix:sgs}.


For text, we manually compose captions describing each of the four accident classes (see Appendix \ref{appendix:caps}), and pair each caption with videos from its respective class to form the training examples. We also experiment with two sets of captions, however, we focus the scope of this work primarily on the scene graph generation process from video frames rather than caption generation. It should be noted that although these captions are used during training for aligning the SGE, they are not used during inference on the final model because they are not from our dataset and provide a one-to-one mapping directly to the accident classes. Through fine-tuning the classification head using these captions, the model can achieve 100\% accuracy. Instead, we fine-tune and test the classification head using the same caption, as noted in Appendix \ref{appendix:caps}, for all examples.

\subsection{Scene Graph Encoder} \label{section:sge}
In stage two, the SGE encodes sequences of scene graphs derived from a given video to fixed-length embeddings. To do this, a multirelational graph convolutional network (MRGCN) as employed in \cite{spatio_temporal_sge} is used, which includes an attention mechanism along with LSTMs to model the spatial and temporal relations of the scene graphs generated for a given video. The SGE may be pre-trained for the classification task before aligning the scene graphs with the language and vision modalities.


\subsection{Multimodal Alignment} \label{section:alignment}
For stage three, we use the CLIP model \cite{clip}, but replace the CLIP image encoder with X-CLIP \cite{xclip}, a pre-trained video encoder, to accept videos rather than images. We leave CLIP's text encoder in place. 
We then freeze the weights from the video and text encoders and align them to the SGE encoder, which takes the new ``scene graph modality'' as input.

\subsection{Fine Tuning for Downstream Task}
The final stage involves training a classification head that accepts embeddings from the three modalities and outputs an accident class for a traffic scene. This requires the selection of a method to fuse the signal from the three modalities followed by training a small network. For this, we experiment with different late fusion techniques such as taking a weighted linear combination of the modality outputs \cite{duality} and training various MLP classifiers \cite{kiela2018efficient} with and without activations on top of the concatenated embeddings. Based on these experiments, we choose the approach of concatenating the vectors from the three modalities as described in \cite{mmclftn} and, following \cite{wu2022wav2clip}, train a 2-layer MLP with ReLU activations as our classification head.

\section{Experimental Design}

\subsection{Dataset}
Our multimodal approach requires data in the form of videos, text, and scene graphs. We focus on using the Detection of Traffic Anomaly (DoTA) dataset \cite{dota} for videos for this task. DoTA is a dataset of 4,677 curated videos of traffic anomalies from YouTube, and was specifically designed to answer the following three questions about each video: (1) When does the anomalous event (i.e., accident) start and end; (2) Where are the anomalous regions of each video frame; and (3) What type of anomaly is taking place. For our use case, question one is irrelevant since we only sample frames from the anomalous portion of the video based on the annotations of the DoTA authors. Question two is also less essential for us since we use a SGG, which extracts the necessary objects from video frames and models their relevant spatial relationships. We therefore prioritize answering the third question, that is, classifying different traffic anomalies. However, given the multitude of classes and the inherent challenge in distinguishing them even with the naked eye, we limit the problem to a subset of four of the available classes. 

In addition to the video classes, \cite{dota} also divides the videos into two types (which are different from classes), \textit{ego} (where the vehicle from which the video recording is captured is involved in the accident), and its opposite, \textit{non-ego}. Every video class has videos of both types (e.g., there are both ego and non-ego videos that fall into the \textit{turning} class). By default, rs2v is designed to only model relationships between the ego vehicle and other objects, not among other objects themselves, so we further restrict our task to the set of ego videos available in DoTA. In total, this leaves us with 4 classes and 2,163 videos. The classes labels, describing the vehicle movement as the accident occurs, are as follows: \textit{moving\_ahead\_or\_waiting}, \textit{oncoming}, \textit{turning}, and \textit{lateral}.

\subsection{Baselines and Metrics}
In order to evaluate the usefulness of representing traffic scenes graphically for classification, we verify whether a SGE alone is superior to randomly guessing an accident class. We then assess the effectiveness of SGs in a multimodal classifier by comparing the difference in performance between using just text and video embeddings versus including the SG modality. Regarding metrics, we focus on accuracy and balanced accuracy. Accuracy is simply defined as the proportion of the correctly classified predictions out of the total number of predictions and is an interpretable metric that can be easily compared across benchmarks. However, there is significant imbalance in the dataset, with the \textit{turning} class appearing over twice as frequently as the other three. We use balanced accuracy, computed as the (unweighted) average recall over all classes, to give equal priority to each accident class.

\subsection{Modality Alignment, Pre-training, and Hyperparameters}
The core \abbv{} model consists of three encoders for three different modalities: text, vision, and graph. For our classification task, a MLP classifier head is placed at the end of the model to output scores for each of the four classes. The text and vision encoders are both kept frozen throughout all of our experiments as we train the SGE to align its embedding space with those of text and vision using Symmetric Cross Entropy Loss \cite{wang2019symmetric}. To evaluate the effectiveness of this training regiment, we analyze the classification performance with and without training for modality alignment. We also experiment with the effect of pre-training the SGE before conducting the alignment training. As an additional pre-training baseline, we use a model provided in \cite{spatio_temporal_sge}, which was pre-trained on a synthetic traffic dataset generated using the CARLA traffic simulator \cite{carla}. Finally, we conduct tests to evaluate the performance of different hyperparameter settings (batch size, training time) for the alignment training.

\subsection{Implementation Details}
We train the SGE for 200 epochs on the default hyperparameter settings as specified in \cite{yu2020scenegraph}. After the SGE pre-training experiments, we select the SGE with the lowest validation loss to use for the modality alignment training. 
During alignment training, we keep most of the default hyperparameter settings, except for varying the batch size (32 to 128) and the number of epochs (8 to 20).

\section{Results and Discussion}
\subsection{SGE Pre-training Results}
As seen in Table \ref{tab:performance}, pre-training the SGE on DoTA allows it to perform classification approximately 13 percentage points better than random classification, which would have an expected accuracy of 25\%. This suggests that spatial and temporal information encoded in the graph embeddings produced by the SGE contain useful information towards accident classification. Unsurprisingly, the model pre-trained on the synthetic CARLA dataset does not achieve as strong performance on our task as the model trained on DoTA. This is understandable, given that the former was trained to classify whether or not a traffic maneuver is deemed to be risky \cite{spatio_temporal_sge} rather than to classify traffic accidents. 
Furthermore, the scenes in CARLA do not vary as much as those in DoTA, and do not contain examples from the four classes we use. 
However, by further training the CARLA model on DoTA, we achieve similar results to training on DoTA alone (compare two rightmost columns of Table \ref{tab:performance}). For the remaining experiments in the paper, when unspecified, we use the SGE pre-trained only on the DoTA dataset.

\begin{table}[htbp]
    \centering
    \caption{SGE Performance with varying pre-train data.\\ Class-balanced accuracy scores are indicated with a (b)}
    \label{tab:performance}
    \resizebox{\columnwidth}{!}{
    \begin{tabular}{lccc}
        \toprule
        \multicolumn{1}{c}{\textbf{Pre-training Data}} & \textbf{CARLA} & \textbf{DoTA} & \textbf{CARLA+DoTA} \\
        \midrule
        Train accuracy          & -     & 62.72 & 64.09 \\
        Train accuracy (b)       & -     & 64.78 & 61.42 \\
        Validation accuracy      & -     & 42.49 & 41.91 \\
        Validation accuracy (b)  & -     & 40.79 & 41.42 \\
        Test accuracy           & 17.32 & 37.88 & 36.03 \\
        Test accuracy (b)        & 26.33 & 38.05 & 37.81 \\
        \bottomrule
    \end{tabular}
    }
\end{table}

\begin{table}[htbp]
    \centering
    \caption{Test Accuracy Comparison between three system settings. In the first setting, the SGE is not used and the inputs to the system are only from the video and text modalities (the SGG and SGE are unused and stage \ref{section:alignment} from the pipeline is skipped). In the second, the SGE is used and aligned with the vision and text encoders before running classification (full architecture is used, but SGE pre-training done in stage \ref{section:sge} may be skipped). In the third, the SGE is used but not aligned with the other encoders (stage \ref{section:alignment} is skipped). All values are obtained using a batch size of 32 and 10 epochs of training}
    \label{tab:test_accuracy}
    \resizebox{\columnwidth}{!}{
    \begin{tabular}{lcccccc}
        \toprule
        \textbf{Train Setting} & \multicolumn{1}{c}{\textbf{No SGE}} & \multicolumn{3}{c}{\textbf{SGE with Alignment}} & \multicolumn{1}{c}{\textbf{SGE without Alignment}} \\
        \cmidrule(lr){1-1} \cmidrule(lr){2-2} \cmidrule(lr){3-5} \cmidrule(lr){6-6}
        Pre-training Data & None & \textcolor{blue}{None$^{\S}$} & CARLA & \textcolor{red}{DoTA$^{*}$} & DoTA \\
        \midrule
        Test accuracy & 57.04 & \textcolor{blue}{54.97} & 52.66 & \textcolor{red}{58.66} & 57.97 \\
        Test accuracy (b) & 53.22 & \textcolor{blue}{53.53} & 53.97 & \textcolor{red}{56.97} & 57.74 \\
        \bottomrule
    \end{tabular}
    }
\end{table}

\subsection{SGE with Alignment Results}\label{subsec:SGE-with-alignment}
The primary classification results after training for alignment can be seen in Table \ref{tab:test_accuracy}. In Column 1, we verify the baseline accuracy of the model when the scene graph modality is not used at all. Looking at the balanced accuracy score, it is interesting to note that the performance on this setting is only slightly under that of the aligned model when no pre-training data is used or if only the CARLA dataset is used (Columns 2, 3). However, pre-training the model on the DoTA dataset (Column 4) shows a further boost in performance by about 3\%. This boost suggests that the SG embeddings may be encoding some information about the traffic scenes that cannot be fully captured by either the language or vision modalities.

\subsubsection{Pre-trained SGE with Alignment}
Regarding the case when the model is pre-trained on the DoTA dataset, the model actually has a slightly better performance without undergoing the alignment training (Column 5) versus with alignment (Column 4). This seems to suggest that the model is not learning further useful information during alignment training. One possible explanation for this is that the data domains of the frozen encoders are not similar enough to DoTA for any positive transfer of data domain knowledge to occur \cite{negtrans}. Another possibility has to do with the alignment training procedure, which is based on CLIP's training algorithm and therefore relies heavily on using a large batch size at each step \cite{clip}. This is because the algorithm performs contrastive training on each of the three pairs of embedding spaces (vision-graph, text-graph, and vision-text) for each batch of traffic accident examples. However, given the limited size of the DoTA dataset and compute constraints, our experiments in Table \ref{tab:test_accuracy} use a batch size of 32 and are limited to 10 epochs of training. We hypothesize that a larger batch size and/or number of training epochs would increase the accuracy gain of the training alignment step. To test this hypothesis, we carried out some additional experiments, the results of which are shown in Table \ref{tab:pre-training_dota}, and indicate that both increasing batch size and the number of epochs give similar results to the ``SGE without Alignment'' case shown in Table \ref{tab:test_accuracy} (Column 5) rather than worse results (Columns 2-4 of Table \ref{tab:test_accuracy}). It is possible that further training of the larger batch sizes would have improved these results beyond the values in Column 5.

\begin{table}[htbp]
    \centering
    \caption{Alignment training with different hyperparameter configurations, using a SGE pre-trained on DoTA. ``e'' is number of epochs, ``bs'' is batch size for the alignment training. Column 4 in Table \ref{tab:test_accuracy} is the same as Column 1 in this table}
    \label{tab:pre-training_dota}
    \resizebox{\columnwidth}{!}{
    \begin{tabular}{lcccc}
        \toprule
        \multicolumn{5}{c}{\textbf{Pre-training on DoTA \& SGE with Alignment}} \\
        \midrule
        & \textcolor{red}{10e, bs32$^{*}$} & 20e, bs32 & 20e, bs64 & 10e, bs128 \\
        \midrule
        Test accuracy & \textcolor{red}{58.66} & 58.2 & 57.51 & 60.51 \\
        Test accuracy (b) & \textcolor{red}{56.97} & 57.77 & 57.53 & 57.72 \\
        \bottomrule
    \end{tabular}
    }
\end{table}

\subsubsection{No Pre-trained SGE with Alignment}
Another surprising point arising from Table \ref{tab:test_accuracy}, which was mentioned at the start of Section \ref{subsec:SGE-with-alignment}, is that there is very little improvement when training the SGE for alignment from scratch (Column 2) over the baseline ``No SGE'' case (Column 1). Similar to the case where we pre-trained on DoTA, it is once again possible that further increasing the batch size could be the solution. This is corroborated by the results in Table \ref{tab:alignment_from_scratch}, where larger batch sizes begin to show a significant increase in performance over the baseline.

\begin{table}[htbp]
    \centering
    \caption{Training for alignment from scratch with different configurations. Column 2 in Table \ref{tab:test_accuracy} is the same as Column 1 in this table}
    \label{tab:alignment_from_scratch}
    \resizebox{\columnwidth}{!}{
    \begin{tabular}{lccc}
        \toprule
        \multicolumn{4}{c}{\textbf{No Pre-training \& SGE with Alignment}} \\
        \midrule
        & \textcolor{blue}{10e, bs32$^{\S}$} & 10e, bs64 & 10e, bs128 \\
        \midrule
        Test accuracy & \textcolor{blue}{54.97} & 57.51 & 59.35 \\
        Test accuracy (b) & \textcolor{blue}{53.53} & 53.94 & 56.00 \\
        \bottomrule
    \end{tabular}
    }
    \vspace{-0.3cm}
\end{table}

\section{Conclusion and Future Work}
\label{sec:conclusion}

In this work, we have shown that encoding traffic information in the form of a scene graph is beneficial towards the goal of accident classification. This is made clear by the pre-training encoder results which show the SGE's ability to beat a random classifier at this task. It was further illustrated that scene graph information can enhance the performance of a vision-language classifier by fusing information from all three modalities. Finally, although we were not able to show an improved score for the classifier after alignment, it should be noted that alignment training using an appropriate hyperparameter configuration does not have a negative effect on performance, and we demonstrated that further training and fine-tuning may improve the score beyond the unaligned case.

There are several areas where future work can be done in this task. Firstly, although three modalities are used, little exploration has been conducted on the language modality. In principle, captions for each video should be derived from the video itself, as the SGs are. This would allow for the captions to be fed into the model to fine-tune the classifier head and run inference with a greater signal coming from the language modality. Next, although they provide some benefit for the model, the generated SGs have a relatively limited structure: only categorical distance relations between a vehicle and objects in its surroundings are generated, along with mappings to a fixed set of three traffic lanes (left, middle, and right). Enabling a semantic extension of the generated SGs may enhance the signal obtained from this modality. Similarly, by modeling relations between all objects in the scene, this pipeline could be expanded for the \textit{non-ego} case as well, and perhaps could be used on more accident classes in the DoTA dataset or similar use cases. Regarding the dataset, although we attempt to resolve the data imbalance by focusing on the balanced accuracy metric, it would be interesting to explore data augmentation techniques as well. Additionally, the MRGCN-based architecture of the SGE was taken directly from \cite{yu2020scenegraph}, but perhaps it can be further improved by modifying either its spatial or temporal modeling components. Given the modular design of our system, another fascinating avenue of exploration would be to incorporate an additional modality (e.g., audio) to include in the alignment step before fusing the embeddings and fine tuning the classification head. Experiments with different modality fusion methods as well as classification heads may also lead to improvements. 

\footnotesize
\bibliography{main}

\begin{thebibliography}{10}
\providecommand{\url}[1]{#1}
\csname url@samestyle\endcsname
\providecommand{\newblock}{\relax}
\providecommand{\bibinfo}[2]{#2}
\providecommand{\BIBentrySTDinterwordspacing}{\spaceskip=0pt\relax}
\providecommand{\BIBentryALTinterwordstretchfactor}{4}
\providecommand{\BIBentryALTinterwordspacing}{\spaceskip=\fontdimen2\font plus
\BIBentryALTinterwordstretchfactor\fontdimen3\font minus \fontdimen4\font\relax}
\providecommand{\BIBforeignlanguage}[2]{{%
\expandafter\ifx\csname l@#1\endcsname\relax
\typeout{** WARNING: IEEEtran.bst: No hyphenation pattern has been}%
\typeout{** loaded for the language `#1'. Using the pattern for}%
\typeout{** the default language instead.}%
\else
\language=\csname l@#1\endcsname
\fi
#2}}
\providecommand{\BIBdecl}{\relax}
\BIBdecl

\bibitem{xu2021sutdtrafficqa}
L.~Xu, H.~Huang, and J.~Liu, ``Sutd-trafficqa: A question answering benchmark and an efficient network for video reasoning over traffic events,'' 2021.

\bibitem{dota}
Y.~Yao, X.~Wang, M.~Xu, Z.~Pu, Y.~Wang, E.~Atkins, and D.~Crandall, ``Dota: unsupervised detection of traffic anomaly in driving videos,'' \emph{IEEE transactions on pattern analysis and machine intelligence}, 2022.

\bibitem{qasemi2023traffic}
E.~Qasemi, J.~M. Francis, and A.~Oltramari, ``Traffic-domain video question answering with automatic captioning,'' \emph{arXiv preprint arXiv:2307.09636}, 2023.

\bibitem{francis2022distribution}
J.~Francis, B.~Chen, W.~Yao, E.~Nyberg, and J.~Oh, ``Distribution-aware goal prediction and conformant model-based planning for safe autonomous driving,'' \emph{arXiv preprint arXiv:2212.08729}, 2022.

\bibitem{huang2024cadre}
P.~Huang, W.~Ding, B.~Stoler, J.~Francis, B.~Chen, and D.~Zhao, ``Cadre: Controllable and diverse generation of safety-critical driving scenarios using real-world trajectories,'' \emph{arXiv preprint arXiv:2403.13208}, 2024.

\bibitem{stoler2024seal}
B.~Stoler, I.~Navarro, J.~Francis, and J.~Oh, ``Seal: Towards safe autonomous driving via skill-enabled adversary learning for closed-loop scenario generation,'' \emph{arXiv preprint arXiv:2409.10320}, 2024.

\bibitem{stoler2024safeshift}
B.~Stoler, I.~Navarro, M.~Jana, S.~Hwang, J.~Francis, and J.~Oh, ``Safeshift: Safety-informed distribution shifts for robust trajectory prediction in autonomous driving,'' in \emph{2024 IEEE Intelligent Vehicles Symposium (IV)}.\hskip 1em plus 0.5em minus 0.4em\relax IEEE, 2024, pp. 1179--1186.

\bibitem{rs2v}
\BIBentryALTinterwordspacing
A.~V. Malawade, S.~Yu, B.~Hsu, H.~Kaeley, A.~Karra, and M.~A.~A. Faruque, ``roadscene2vec: {A} tool for extracting and embedding road scene-graphs,'' \emph{CoRR}, vol. abs/2109.01183, 2021. [Online]. Available: \url{https://arxiv.org/abs/2109.01183}
\BIBentrySTDinterwordspacing

\bibitem{spatio_temporal_sge}
A.~V. Malawade, S.-Y. Yu, B.~Hsu, D.~Muthirayan, P.~P. Khargonekar, and M.~A.~A. Faruque, ``Spatiotemporal scene-graph embedding for autonomous vehicle collision prediction,'' \emph{IEEE Internet of Things Journal}, vol.~9, no.~12, pp. 9379--9388, 2022.

\bibitem{clip}
\BIBentryALTinterwordspacing
A.~Radford, J.~W. Kim, C.~Hallacy, A.~Ramesh, G.~Goh, S.~Agarwal, G.~Sastry, A.~Askell, P.~Mishkin, J.~Clark, G.~Krueger, and I.~Sutskever, ``Learning transferable visual models from natural language supervision,'' \emph{CoRR}, vol. abs/2103.00020, 2021. [Online]. Available: \url{https://arxiv.org/abs/2103.00020}
\BIBentrySTDinterwordspacing

\bibitem{xclip}
B.~Ni, H.~Peng, M.~Chen, S.~Zhang, G.~Meng, J.~Fu, S.~Xiang, and H.~Ling, ``Expanding language-image pretrained models for general video recognition,'' 2022.

\bibitem{yu2020scenegraph}
S.-Y. Yu, A.~V. Malawade, D.~Muthirayan, P.~P. Khargonekar, and M.~A.~A. Faruque, ``Scene-graph augmented data-driven risk assessment of autonomous vehicle decisions,'' 2020.

\bibitem{semsgg}
\BIBentryALTinterwordspacing
M.~Zipfl and J.~M. Z{\"{o}}llner, ``Towards traffic scene description: The semantic scene graph,'' \emph{CoRR}, vol. abs/2111.10196, 2021. [Online]. Available: \url{https://arxiv.org/abs/2111.10196}
\BIBentrySTDinterwordspacing

\bibitem{kgg}
\BIBentryALTinterwordspacing
Y.~Guo, F.~Yin, X.~Li, X.~Yan, T.~Xue, S.~Mei, and C.~Liu, ``Visual traffic knowledge graph generation from scene images,'' in \emph{2023 IEEE/CVF International Conference on Computer Vision (ICCV)}.\hskip 1em plus 0.5em minus 0.4em\relax Los Alamitos, CA, USA: IEEE Computer Society, oct 2023, pp. 21\,547--21\,556. [Online]. Available: \url{https://doi.ieeecomputersociety.org/10.1109/ICCV51070.2023.01975}
\BIBentrySTDinterwordspacing

\bibitem{sgg_in_ad}
\BIBentryALTinterwordspacing
P.~E.~I. Dimasi, ``Scene graph generation in autonomous driving: a neuro-symbolic approach,'' Master's thesis, Politecnico di Torino, 2023. [Online]. Available: \url{http://webthesis.biblio.polito.it/id/eprint/29354}
\BIBentrySTDinterwordspacing

\bibitem{reltr}
\BIBentryALTinterwordspacing
Y.~Cong, M.~Y. Yang, and B.~Rosenhahn, ``Reltr: Relation transformer for scene graph generation,'' \emph{CoRR}, vol. abs/2201.11460, 2022. [Online]. Available: \url{https://arxiv.org/abs/2201.11460}
\BIBentrySTDinterwordspacing

\bibitem{regat}
\BIBentryALTinterwordspacing
L.~Li, Z.~Gan, Y.~Cheng, and J.~Liu, ``Relation-aware graph attention network for visual question answering,'' 2019. [Online]. Available: \url{https://arxiv.org/abs/1903.12314}
\BIBentrySTDinterwordspacing

\bibitem{DG-PGNN}
\BIBentryALTinterwordspacing
M.~Khademi and O.~Schulte, ``Deep generative probabilistic graph neural networks for scene graph generation,'' \emph{Proceedings of the AAAI Conference on Artificial Intelligence}, vol.~34, no.~07, pp. 11\,237--11\,245, Apr. 2020. [Online]. Available: \url{https://ojs.aaai.org/index.php/AAAI/article/view/6783}
\BIBentrySTDinterwordspacing

\bibitem{sayplan}
\BIBentryALTinterwordspacing
K.~Rana, J.~Haviland, S.~Garg, J.~Abou-Chakra, I.~Reid, and N.~Suenderhauf, ``Sayplan: Grounding large language models using 3d scene graphs for scalable robot task planning,'' 2023. [Online]. Available: \url{https://arxiv.org/abs/2307.06135}
\BIBentrySTDinterwordspacing

\bibitem{saxena2024grapheqa}
S.~Saxena, B.~Buchanan, C.~Paxton, B.~Chen, N.~Vaskevicius, L.~Palmieri, J.~Francis, and O.~Kroemer, ``Grapheqa: Using 3d semantic scene graphs for real-time embodied question answering,'' \emph{arXiv preprint arXiv:2412.14480}, 2024.

\bibitem{leveragingsgs}
\BIBentryALTinterwordspacing
C.~Zhang, W.-L. Chao, and D.~Xuan, ``An empirical study on leveraging scene graphs for visual question answering,'' 2019. [Online]. Available: \url{https://arxiv.org/abs/1907.12133}
\BIBentrySTDinterwordspacing

\bibitem{nag2023unbiased}
S.~Nag, K.~Min, S.~Tripathi, and A.~K.~R. Chowdhury, ``Unbiased scene graph generation in videos,'' 2023.

\bibitem{ag}
\BIBentryALTinterwordspacing
J.~Ji, R.~Krishna, L.~Fei{-}Fei, and J.~C. Niebles, ``Action genome: Actions as composition of spatio-temporal scene graphs,'' \emph{CoRR}, vol. abs/1912.06992, 2019. [Online]. Available: \url{http://arxiv.org/abs/1912.06992}
\BIBentrySTDinterwordspacing

\bibitem{wu2022wav2clip}
H.-H. Wu, P.~Seetharaman, K.~Kumar, and J.~P. Bello, ``Wav2clip: Learning robust audio representations from clip,'' 2022.

\bibitem{tatiya2024mosaic}
G.~Tatiya, J.~Francis, H.-H. Wu, Y.~Bisk, and J.~Sinapov, ``Mosaic: Learning unified multi-sensory object property representations for robot learning via interactive perception,'' 2024.

\bibitem{koch2023lang3dsg}
S.~Koch, P.~Hermosilla, N.~Vaskevicius, M.~Colosi, and T.~Ropinski, ``Lang3dsg: Language-based contrastive pre-training for 3d scene graph prediction,'' 2023.

\bibitem{huang2023structureclip}
Y.~Huang, J.~Tang, Z.~Chen, R.~Zhang, X.~Zhang, W.~Chen, Z.~Zhao, Z.~Zhao, T.~Lv, Z.~Hu, and W.~Zhang, ``Structure-clip: Towards scene graph knowledge to enhance multi-modal structured representations,'' 2023.

\bibitem{mmclftn}
\BIBentryALTinterwordspacing
W.~C. Sleeman, R.~Kapoor, and P.~Ghosh, ``Multimodal classification: Current landscape, taxonomy and future directions,'' \emph{ACM Comput. Surv.}, vol.~55, no.~7, dec 2022. [Online]. Available: \url{https://doi.org/10.1145/3543848}
\BIBentrySTDinterwordspacing

\bibitem{mmanal}
\BIBentryALTinterwordspacing
M.~Pawłowski, A.~Wróblewska, and S.~Sysko-Romańczuk, ``Effective techniques for multimodal data fusion: A comparative analysis,'' \emph{Sensors}, vol.~23, no.~5, 2023. [Online]. Available: \url{https://www.mdpi.com/1424-8220/23/5/2381}
\BIBentrySTDinterwordspacing

\bibitem{mmsurvey}
\BIBentryALTinterwordspacing
J.~Gao, P.~Li, Z.~Chen, and J.~Zhang, ``{A Survey on Deep Learning for Multimodal Data Fusion},'' \emph{Neural Computation}, vol.~32, no.~5, pp. 829--864, 05 2020. [Online]. Available: \url{https://doi.org/10.1162/neco\_a\_01273}
\BIBentrySTDinterwordspacing

\bibitem{duality}
L.~Kaliciak, H.~Myrhaug, A.~Goker, and D.~Song, ``On the duality of specific early and late fusion strategies,'' in \emph{17th International Conference on Information Fusion (FUSION)}, 2014, pp. 1--8.

\bibitem{kiela2018efficient}
D.~Kiela, E.~Grave, A.~Joulin, and T.~Mikolov, ``Efficient large-scale multi-modal classification,'' 2018.

\bibitem{wang2019symmetric}
Y.~Wang, X.~Ma, Z.~Chen, Y.~Luo, J.~Yi, and J.~Bailey, ``Symmetric cross entropy for robust learning with noisy labels,'' 2019.

\bibitem{carla}
A.~Dosovitskiy, G.~Ros, F.~Codevilla, A.~Lopez, and V.~Koltun, ``Carla: An open urban driving simulator,'' 2017.

\bibitem{negtrans}
W.~Zhang, L.~Deng, L.~Zhang, and D.~Wu, ``A survey on negative transfer,'' \emph{IEEE/CAA Journal of Automatica Sinica}, vol.~10, no.~2, pp. 305--329, 2023.

\end{thebibliography}
\bibliographystyle{IEEEtran}
\clearpage


\begin{appendices}

\section{Generated Scene Graph Examples}
\label{appendix:sgs}


\begin{figure}[!h]
    \centering
    \includegraphics[width=1\linewidth]{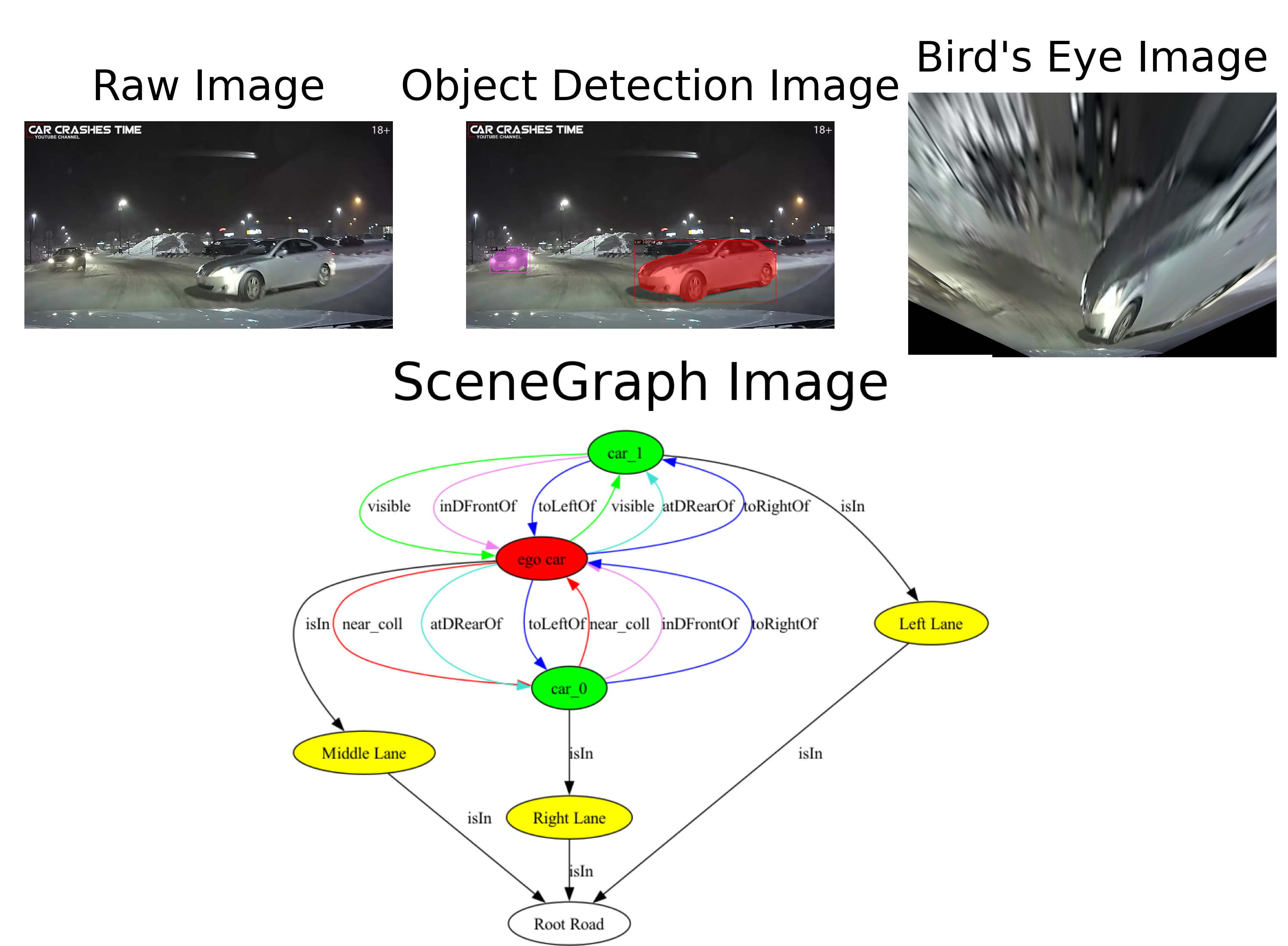}
    \caption{A scene graph generated using the rs2v \cite{rs2v} tool. Starting from a video frame (Raw Image), the SGG first detects objects in the scene (Object Detection Image) and generates the BEV (Bird's Eye Image) before creating the scene graph representation (SceneGraph Image). This scene graph shows the ego car relative to two other vehicles, one categorized as in the left lane, the other right. The closer vehicle is recognized as being near collision (with the edge attribute ``near\_coll''), whereas the farther vehicle is registered in the scene graph as simply being ``visible''.}
    \label{fig:gensgs}
\end{figure}

\begin{figure}[!h]
    \centering
    \includegraphics[width=1\linewidth]{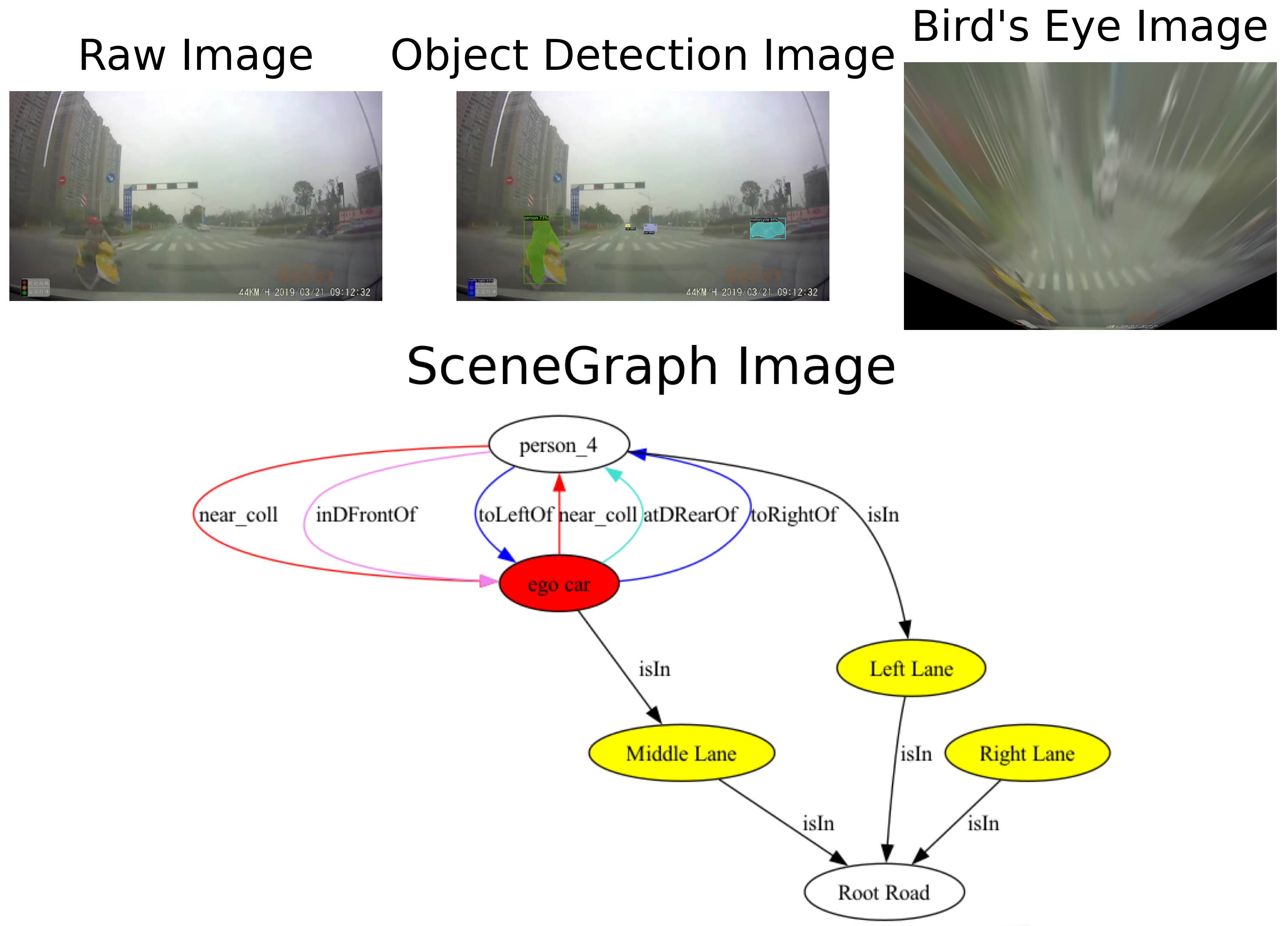}
    \caption{Starting from a video frame, the SGG first detects objects in the scene and generates the BEV image before creating the scene graph representation. This scene graph shows the ego car relative to a person on a scooter, categorized as in the left lane. The person is recognized as being near collision, with the edge attribute ``near\_coll''. Note that there is a vehicle detected in the distance (right part of detection image, highlighted in blue) but it is not included in the graph since it is too far.}
    \label{fig:gensgs2}
\end{figure}

\vspace{3cm}
\section{Captions}
\label{appendix:caps}

\begin{table}[h]
    \centering
    \caption{Two sets of captions that were experimented with for training during alignment. All results shown in this report used Style B. For fine-tuning the classification head, all examples were trained with the same generic caption, ``An accident as a result of a vehicle doing something.''}
    \begin{tabular}{|p{0.24\linewidth}|p{0.32\linewidth}|p{0.32\linewidth}|}
    \hline
    \textbf{Accident Class} & \textbf{Caption Style A} & \textbf{Caption Style B} \\ \hline
    Moving Ahead or Waiting & The vehicle is moving ahead or waiting in the accident. & An accident as a result of a vehicle moving into another vehicle. \\ \hline
    Oncoming & The vehicle is hitting an oncoming vehicle in the accident. & An accident as a result of a vehicle hitting an oncoming vehicle. \\ \hline
    Turning & The vehicle is turning in the accident. & An accident as a result of a vehicle turning. \\ \hline
    Lateral & The vehicle is moving laterally in the accident. & An accident as a result of a vehicle moving laterally. \\ \hline
    \end{tabular}
    \label{tab:accident_scenarios}
\end{table}

\end{appendices}

\end{document}